# Building Safe and Deployable Clinical Natural Language Processing under Temporal Leakage Constraints

Ha Na Cho
Informatics, Computer Science
University of California Irvine
Irvine, USA

Sairam Sutari
Informatics, Computer Science
University of California Irvine
Irvine, USA

Alexander Lopez
Department of Neurosurgery
University of California Irvine
Irvine, USA

Hansen Bow
Department of Neurosurgery
University of California Irvine
Irvine, USA

Kai Zheng
Informatics, Computer Science
University of California Irvine
Irvine, USA

*Abstract*—Clinical natural language processing (NLP) models have shown promise for supporting hospital discharge planning by leveraging narrative clinical documentation. However, note-based models are particularly vulnerable to temporal and lexical leakage, where documentation artifacts encode future clinical decisions and inflate apparent predictive performance. Such behavior poses substantial risks for real-world deployment, where overconfident or temporally invalid predictions can disrupt clinical workflows and compromise patient safety. This study focuses on system-level design choices required to build safe and deployable clinical NLP under temporal leakage constraints. We present a lightweight auditing pipeline that integrates interpretability into the model development process to identify and suppress leakage-prone signals prior to final training. Using next-day discharge prediction after elective spine surgery as a case study, we evaluate how auditing affects predictive behavior, calibration, and safety-relevant trade-offs. Results show that audited models exhibit more conservative and better-calibrated probability estimates, with reduced reliance on discharge-related lexical cues. These findings emphasize that deployment-ready clinical NLP systems should prioritize temporal validity, calibration, and behavioral robustness over optimistic performance.

## I. Introduction

Timely and accurate hospital discharge planning plays a critical role in bed management, staffing coordination, and postoperative care delivery [1]. Predictive models that forecast near-term discharge have the potential to support proactive clinical decision-making and improve operational efficiency [1], [2]. In recent years, clinical natural language processing (NLP) has been increasingly applied to this task, motivated by the rich contextual information contained in progress notes and other narrative documentation [3]. Despite encouraging performance reports, the deployment of note-based clinical NLP systems remains limited.

A key challenge lies in temporal leakage. Clinical documentation often reflects downstream decisions, care plans, or anticipated outcomes, especially in the peri-discharge period. As a result, NLP models may learn spurious lexical shortcuts that encode future information, even when strict timestamp-based filtering is applied. Models trained under these conditions can appear highly accurate while relying on signals that are temporally invalid at the point of prediction [4], [5]. In operational settings, such behavior undermines trust, inflates confidence, and risks unsafe recommendations. Existing mitigation strategies [4], [5] primarily rely on heuristic preprocessing, including exclusion of discharge summaries, removal of explicit discharge terminology, or coarse temporal alignment of notes. While these approaches address overt leakage, they do not systematically capture indirect proxy signals or documentation artifacts that models exploit during training. Moreover, most interpretability methods in clinical NLP are applied after model development, serving a descriptive role that explains predictions without constraining what the model is allowed to learn [6], [7].

In this work, we argue that building deployable clinical NLP systems requires a shift in emphasis from performance optimization to behavioral validity. We focus on three design priorities: enforcing temporal validity, promoting well-calibrated and conservative predictive behavior, and ensuring feasibility in resource-constrained clinical environments. To this end, we introduce a lightweight auditing pipeline that integrates interpretability into the training process as a control mechanism. By identifying and suppressing leakage-prone signals prior to final model fitting, the proposed approach constrains the hypothesis space of clinical NLP models and encourages safer predictive behavior. Using next-day discharge prediction following elective spine surgery as a representative use case, we demonstrate how interpretability-guided auditing influences model behavior beyond standard discrimination metrics. Our results highlight the importance of calibration and robustness in deployment-facing clinical NLP systems and illustrate how lightweight, retrainable pipelines can support safer integration of machine learning into real-world healthcare workflows.

## II. Related Work

### A. Clinical Natural Language Processing and Discharge Prediction

Early work [8], [9] on hospital discharge and length-of-stay prediction primarily relied on structured electronic health record (EHR) data, including demographics, diagnoses, procedures, and laboratory measurements. Classical machine learning models trained on such variables demonstrated reasonable discrimination and stability, particularly in surgical populations, and remain widely used in operational settings.

These approaches benefit from well-defined temporal alignment and relatively low risk of information leakage.

More recent studies [10], [11] have incorporated unstructured clinical notes to capture contextual information not represented in structured fields. Transformer-based models [12], [13] and large language models have been applied to discharge prediction and related postoperative outcomes, often reporting improved predictive performance. However, these gains are frequently evaluated under retrospective settings and focus on discrimination metrics, with limited examination of whether learned signals are temporally valid at the point of prediction. As a result, strong reported performance may reflect exploitation of documentation artifacts rather than clinically meaningful inference.

From a deployment perspective, reliance on unstructured text introduces additional risks. Clinical notes are authored as part of ongoing care processes and may implicitly encode downstream decisions, care plans, or anticipated discharge timing. Without careful control, note-based models can learn shortcuts that undermine their utility in real-world decision support.

*B. Temporal Leakage in Clinical Prediction Models*

Temporal leakage is a well-recognized challenge in clinical machine learning, particularly for tasks involving future outcomes such as discharge timing, readmission, or deterioration [14]. Leakage can arise when information recorded after the prediction window is inadvertently included in model inputs or when proxy variables encode future events. In clinical text, this issue is especially acute, as narrative documentation may reflect anticipated actions or decisions before they are formally executed. Common mitigation strategies include timestamp-based filtering, exclusion of discharge summaries, and rule-based removal of explicit outcome-related terms [15]. While these heuristics reduce overt leakage, they do not systematically address indirect proxy signals or documentation practices that vary across clinicians and institutions. As a result, models may continue to rely on temporally invalid cues even under strict preprocessing rules. This gap highlights the need for approaches that go beyond static filtering and directly assess how models use available inputs during training.

*C. Interpretability and Model Auditing in Clinical Machine Learning*

Interpretability has become a central concern in clinical machine learning due to the high-stakes nature of medical decision-making [16]. Post-hoc explanation methods, including feature attribution and saliency-based approaches, are widely used to inspect trained models, support qualitative validation, and increase clinician trust [17]. In clinical NLP, such methods are typically applied after model development to rationalize predictions or identify influential tokens. However, post-hoc explanations alone do not prevent models from learning spurious or leakage-prone dependencies during training. Recent work [18] has emphasized the need for model auditing and governance mechanisms that assess robustness, fairness, and validity prior to deployment. Despite this growing awareness, interpretability is still most often treated as a descriptive tool rather than as an active component of the modeling pipeline.

In contrast, system-oriented perspectives [19] in healthcare AI argue that deployment readiness depends on controlling model behavior, not only explaining it. This includes enforcing temporal validity, promoting calibrated and conservative predictions, and ensuring that models operate within clinically acceptable boundaries. Lightweight, repeatable auditing mechanisms are particularly important in resource-constrained environments where frequent retraining and monitoring are required.

This work builds on prior efforts in discharge prediction, temporal leakage mitigation, and clinical interpretability, while shifting the focus toward deployment-oriented system design. Rather than proposing a new predictive architecture or benchmarking performance across many models, we focus on how interpretability can be integrated into the training process as a control mechanism. By identifying and suppressing leakage-prone signals before final model fitting, the proposed approach addresses temporal validity at the system level and supports safer deployment of clinical NLP models.

III. METHODS

*A. Study Design, Cohort, and Outcome Definition*

We conducted a retrospective modeling study to examine how temporal and lexical leakage affects the behavior and deployability of clinical NLP models for discharge prediction (Fig. 1). The prediction task was defined as next-day hospital discharge following elective spine surgery. The binary outcome variable indicated whether a patient was discharged on the calendar day immediately following surgery (next-day discharge=1) or remained hospitalized beyond that point (next-day discharge=0). To ensure temporal correctness at the prediction point, all input features were constrained to information available prior to the discharge decision window. Clinical notes were aligned relative to the index surgical date. For patients discharged on postoperative day one, only notes authored on the day of surgery (postoperative day 0) were included. For patients with longer hospitalizations, notes authored on or after the day preceding discharge were excluded.

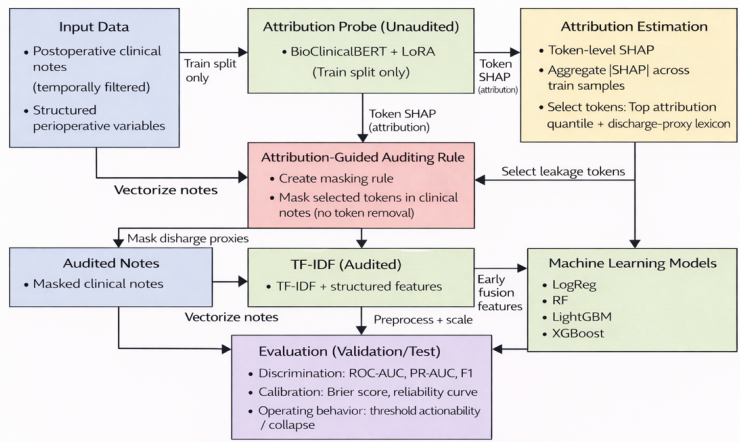

Fig. 1. Overview of the proposed interpretability-guided auditing pipeline for temporally valid clinical natural language processing.

The study cohort consisted of adult patients (age > 18) who underwent elective spine surgery at a single academic medical center. De-identified EHR data were extracted under institutional review board approval. The dataset included unstructured postoperative clinical notes authored during the eligible prediction window, structured perioperative clinical variables, and next-day discharge outcomes. After applying temporal alignment and eligibility criteria and performing patient-level merging across modalities, the final cohort contained 1,251 unique patients. All analyses in this study were conducted using this fixed cohort to ensure consistent patient counts across structured features, note representations, and fused modeling inputs.

*B. Data Processing and Feature Construction*

To prevent patient-level leakage and preserve outcome prevalence, we performed stratified patient-level partitioning into training, validation, and test sets using a fixed random seed with a 64/16/20 split. This yielded fused feature matrices of shape (1251, 10627), (313, 10627), and (392, 10627), respectively. All discrimination, calibration, and confusion-matrix results were computed on the validation or test partitions only. Attribution-based auditing rules and masking patterns were derived exclusively from the training split. Structured covariates were processed using a standardized pipeline. Boolean variables were cast to integer format, and numeric columns were coerced to consistent numeric types. Missing values were imputed using mean imputation computed on the training split, and the same imputation parameters were applied to validation and test data. All structured features were then standardized using z-score scaling based on training-set statistics to ensure consistent feature scaling across splits. After preprocessing, the structured feature block contained 627 dimensions.

Clinical notes were first normalized to reduce encoding-related artifacts. Text was Unicode-normalized (NFKD) and converted to ASCII to remove non-standard characters introduced during extraction. We then applied a rule-based masking step to suppress explicit discharge-proxy terminology and short-horizon timing phrases that can encode downstream clinical decisions. Discharge-related stems (e.g., discharge, dispo, dc) and timing expressions (e.g., "next day", "within 24 hours", "by morning", "overnight") were replaced with a mask token. In addition, occurrences of "tomorrow" were masked only when appearing in discharge-planning contexts (e.g., co-occurring with "discharge", "dc", "go home", "leave"), to avoid removing clinically neutral temporal references. Notes were then represented using a TF-IDF vectorizer fit on training notes only (unigrams/bigrams, max features=10,000, min df=3), yielding a 10,000-dimensional text feature block.

*C. Attribution-Guided Auditing*

To characterize how a standard note-based transformer behaves under timestamp-based filtering, we fine-tuned a domain-specific transformer encoder (Bio_ClinicalBERT) as an attribution probe. To support parameter-efficient retraining and reflect resource-constrained clinical environments, we applied Low-Rank Adaptation (LoRA) to the self-attention layers. Specifically, LoRA adapters were applied to the query and value projection modules (r = 8, alpha = 16, and dropout = 0.1), while the base BioClinicalBERT parameters were frozen and only adapter parameters were updated. The model was trained as a binary sequence classifier with cross-entropy loss and class weighting to address outcome imbalance. Optimization used AdamW (learning rate of 2e-5, batch size of 8) for 3 training epochs. Inputs were tokenized using the BioClinicalBERT tokenizer and padded or truncated to a maximum sequence length of 512 tokens. No discharge-term filtering was applied at this stage, allowing the model to learn freely from temporally eligible text. This model was used to surface leakage-prone lexical dependencies for auditing.

Attribution-guided auditing was implemented as a training-time governance step to suppress leakage-prone discharge-proxy language before training downstream fusion classifiers. Token attributions were computed on the training split only using SHAP, and tokens were selected for masking based on (i) extreme attribution magnitude (quantile=0.99) and (ii) overlap with a predefined discharge-proxy lexicon. Selected terms were masked using case-insensitive, word-boundary replacement with the model mask token. Auditing was applied once, and the masked notes were used to construct the audited TF-IDF representation for early fusion.

*D. Downstream Models, Calibration, and Evaluation*

We constructed an early-fusion representation by concatenating the scaled structured feature block and the TF-IDF note block, resulting in 10,627 fused features per patient. Using this shared early-fusion feature space, we trained four retrainable deployment-oriented baselines (logistic regression, random forest, LightGBM, and XGBoost) under identical train/validation/test splits with class-imbalance handling and fixed random seeds. We further evaluated calibrated fusion variants and two ensemble strategies on the validation partition using calibrated probabilistic outputs from selected base learners (logistic regression, LightGBM, and XGBoost). Random forest was evaluated as a tree-based early-fusion baseline but was not included in the calibration and ensemble experiments, since gradient-boosted tree models (LightGBM and XGBoost) showed stronger discrimination and more stable probabilistic behavior in preliminary comparisons.

For voting, we applied soft voting by averaging predicted probabilities across models before thresholding. For stacking, we trained a logistic regression meta-learner using base-model probability outputs as inputs to generate final predictions. These ensemble variants were included to examine deployment-relevant precision-recall trade-offs under different aggregation policies.

Probability calibration was assessed using reliability curves and the Brier score. When calibration was applied, we used sigmoid calibration fit on the validation set to map raw model scores to calibrated probabilities. Reliability curves were constructed using 10 probability bins by comparing mean predicted probability with the empirical event rate per bin. Calibrated probabilities were used only for evaluation on held-out partitions and were not used to modify model training.

Model evaluation emphasized both discrimination and calibration to reflect deployment-relevant behavior. Discrimination was assessed using ROC-AUC and precision, recall, and F1-score for the next-day discharge class. Calibration was assessed using the Brier score and reliability curves. In addition to quantitative metrics, qualitative attribution analysis was conducted using token-level SHAP explanations to examine whether auditing suppressed discharge-proxy lexical cues and shifted attribution toward broader postoperative context. All attribution-guided masking rules were derived from training data only, and all reported results were computed on validation or test partitions excluded from both model fitting and auditing rule construction.

Ablation settings were evaluated to isolate the contribution of each modality under identical temporal filtering and splits. We compared three configurations: (1) Structured-only (LightGBM) using perioperative structured variables, (2) Unstructured-only (Unaudited BioClinicalBERT) using temporally eligible notes without lexical suppression, and (3) Unstructured-only (Audited BioClinicalBERT) using the same note inputs after attribution-guided lexical masking.

## IV. Results

### A. Early Fusion Baseline Performance

Table I summarizes the early fusion baselines trained on concatenated structured and notes. Under the controlled temporal filtering rules described in Methods, gradient boosting approaches produced the strongest overall discrimination and the most favorable minority-class performance on the validation partition. In particular, the LightGBM fusion baseline achieved a well-balanced operating point at the default 0.5 threshold, indicating that combining perioperative structured variables with temporally valid note representations can support actionable next-day discharge detection. Across the four baseline classifiers, LightGBM also exhibited the strongest calibration profile among the non-calibrated models, with comparatively lower probability error as reflected by the Brier score.

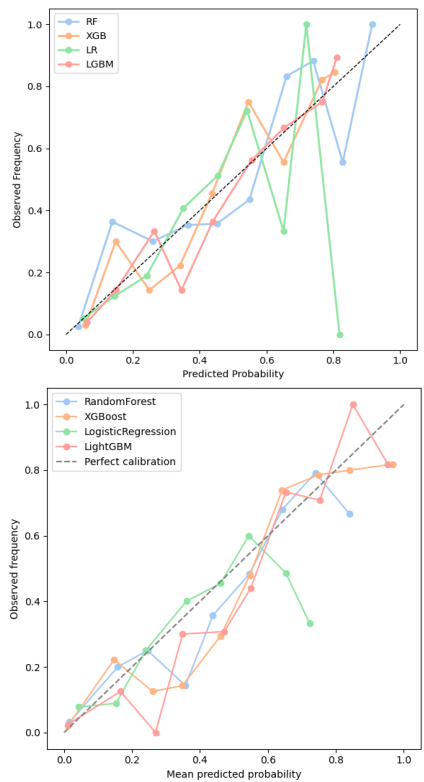

Fig. 2. Calibration curves for baseline early-fusion models (top) and attribution-guided lexical masking-guided fusion models (bottom) on validation set. The dashed line denotes perfect calibration.

### B. Calibrated Fusion Models and Ensembles

To assess whether probability validity could be improved while preserving actionable detection of next-day discharge events, we evaluated calibrated fusion variants and ensemble strategies on the validation partition (Table II). Calibrated LightGBM and calibrated XGBoost achieved comparable

TABLE I. Baseline early-fusion performance of deployment-friendly classifiers (Structured + TF-IDF, validation set)

| Model | Acc | P_0 | R_0 | F1_0 | P_1 | R_1 | F1_1 | ROC_AUC | Brier |
|---|---|---|---|---|---|---|---|---|---|
| LogisticRegression | 0.76 | 0.90 | 0.77 | 0.83 | 0.48 | 0.71 | 0.57 | 0.78 | 0.19 |
| RandomForest | 0.85 | 0.90 | 0.89 | 0.89 | 0.64 | 0.67 | 0.65 | 0.90 | 0.10 |
| XGBoost | 0.86 | 0.93 | 0.88 | 0.90 | 0.65 | 0.78 | 0.71 | 0.92 | 0.10 |
| LightGBM | 0.87 | 0.94 | 0.88 | 0.91 | 0.67 | 0.82 | 0.74 | 0.92 | 0.09 |

a. Acc denotes overall accuracy. P_0/R_0/F1_0 and P_1/R_1/F1_1 denote precision, recall, and F1-score for the negative class (0) and positive class (1; next-day discharge), respectively.

TABLE II. Calibrated fusion models and ensemble variants on the validation set

| Model | Acc | P_0 | R_0 | F1_0 | P_1 | R_1 | F1_1 | ROC_AUC | Brier |
|---|---|---|---|---|---|---|---|---|---|
| Logistic Regression Calibrated | 0.78 | 0.81 | 0.93 | 0.86 | 0.52 | 0.27 | 0.35 | 0.79 | 0.14 |
| XGBoost Calibrated | 0.88 | 0.93 | 0.92 | 0.92 | 0.74 | 0.75 | 0.75 | 0.93 | 0.08 |
| LightGBM Calibrated | 0.88 | 0.93 | 0.92 | 0.92 | 0.73 | 0.77 | 0.75 | 0.93 | 0.08 |
| Voting (LR,LGBM,XGB) | 0.88 | 0.91 | 0.93 | 0.92 | 0.76 | 0.71 | 0.74 | 0.91 | 0.08 |
| Stacking (Meta-LR; LR,LGBM,XGB) | 0.86 | 0.95 | 0.87 | 0.91 | 0.66 | 0.84 | 0.74 | 0.90 | 0.10 |

TABLE III. PERFORMANCE UNDER ATTRIBUTION-GUIDED LEXICAL MASKING IN EARLY FUSION (VALIDATION)

| Model | P_0 | R_0 | F1_0 | P_1 | R_1 | F1_1 | ROC_AUC | Brier |
|---|---|---|---|---|---|---|---|---|
| LogisticRegression | 0.96 | 0.99 | 0.98 | 0.69 | 0.09 | 0.17 | 0.93 | 0.02 |
| RandomForest | 0.96 | 1.00 | 0.98 | 0.00 | 0.00 | 0.00 | 0.89 | 0.03 |
| XGBoost | 0.96 | 1.00 | 0.98 | 0.00 | 0.00 | 0.00 | 0.91 | 0.02 |
| LightGBM | 0.96 | 1.00 | 0.98 | 0.00 | 0.00 | 0.00 | 0.91 | 0.02 |

TABLE III. DEEP FUSION BIOCLINICALBERT FINE-TUNING RESULTS ON THE TEST SET

| Model | Accuracy | Macro-F1 | Precision_1 | Recall_1 | F1_1 | Support_0 | Support_1 |
|---|---|---|---|---|---|---|---|
| Fusion fine-tuned BioClinicalBERT | 0.78 | 0.44 | 0.00 | 0.00 | 0.00 | 304 | 88 |

b. Macro-F1 is the unweighted mean of the F1-scores for class 0 and 1 (0 = non-next-day discharge, 1 = next-day discharge). Support_0 and Support_1 denote the number of samples in each class.

accuracy and ROC-AUC, while yielding more favorable precision-recall behavior for the discharge class than calibrated logistic regression. Ensemble approaches exhibited distinct trade-offs consistent with deployment decision policies. The voting ensemble increased discharge precision but reduced recall, reflecting a more conservative operating regime that limits false-positive discharge recommendations at the cost of missed discharges. In contrast, stacking increased recall but reduced precision and worsened the Brier score, indicating higher sensitivity coupled with a calibration penalty. These results show that calibration-sensitive metrics are necessary for deployment-facing assessment and strong discrimination can be retained while improving probability validity, but more aggressive sensitivity gains may introduce instability in probabilistic reliability that should be interpreted together with operational decision support constraints. In addition to summary calibration error (Brier score), reliability curves provide a more granular view of probability validity across the score range. As shown in Fig. 2, baseline early-fusion models show noticeable deviations from the diagonal reference line, particularly in the mid-probability range where operational decision thresholds are typically applied. After calibration, LightGBM and XGBoost exhibit improved agreement with the ideal calibration line, indicating more reliable probability estimates. In contrast, ensemble strategies display more irregular reliability patterns, consistent with their higher Brier scores and the trade-offs observed in Table II.

### C. Attribution-guided lexical masking

We next compared the baseline early-fusion setting against an attribution-guided fusion variant in which discharge-proxy terms were suppressed in the clinical note stream using an attribution-informed auditing mask. In the baseline fusion setting, the LightGBM classifier achieved a balanced operating point at the default 0.5 threshold, with the following confusion matrix: true negatives (TN)=226, false positives (FP)=17, false negatives (FN)=19, and true positives (TP)=51 (positive class = next-day discharge. This pattern reflects meaningful detection of next-day discharge cases while maintaining a moderate false-positive rate, supporting clinical plausibility for discharge planning use where both sensitivity and false-positive burden affect workflow.

After applying atttribution-guided lexical suppression, the downstream fusion LightGBM model collapsed to predicting only the majority class at the same threshold (Table III). This collapse represents a critical behavioral outcome. The high ROC-AUC indicates that the model preserved meaningful rank-order discrimination, implying that the fused representation retained some separation between discharge and non-discharge cases. However, the disappearance of precision and recall for the discharge class indicates that the default operating point became maximally conservative. From a deployment perspective, this pattern reveals a strict governance effect. False-positive discharge recommendations are minimized, but operational usefulness is lost unless the decision policy is adapted. Restoring actionable discharge support would require explicit threshold selection or risk-binning rules, refinement of the audit mask to avoid suppressing legitimate discharge readiness signals embedded in clinical context, or cost-sensitive training strategies that preserve minority-class sensitivity under governance constraints. To assess whether this collapse was specific to LightGBM, we evaluated multiple downstream classifiers in the audited fusion setting (Table III). Random Forest, XGBoost, and LightGBM consistently exhibited majority-only prediction behavior under the audited representation, while logistic regression preserved limited positive predictions but with very low recall. This reproducible pattern indicates that audit-driven lexical governance can induce a systematic transition toward conservative behavior across multiple classifier families even when rank-order discrimination remains non-trivial.

### D. Deep Fusion Fine-tuning

We observed an analogous failure mode in a deep fusion fine-tuning setting where BioClinicalBERT was trained to directly predict discharge outcomes in an early-fusion configuration. On the test set, the model predicted only the majority class (Table IV). This behavior is reported as a clinically relevant failure mode because it mirrors the audited fusion collapse. The system becomes maximally conservative at the reporting threshold, achieving high majority-class performance while losing actionable sensitivity for discharge. The result reinforces that under constrained signal conditions, including strong leakage suppression or difficult class imbalance, models can preserve overall accuracy yet fail as decision-support tools for the outcome of interest.

### E. Attribution Evidence and Auditing Effects

Token-level attribution analysis of the unaudited BioClinicalBERT attribution probe revealed systematic dependence on discharge-related lexical cues. Tokens explicitly referencing discharge status or imminent transition of care repeatedly emerged as top contributors to positive discharge predictions, despite strict temporal note filtering. This finding indicates that timestamp alignment alone cannot fully prevent temporally invalid information from being implicitly encoded in eligible clinical documentation. Aggregated attribution patterns suggest that the model learned a structural shortcut tied to documentation practices, instead of relying predominantly on physiological recovery descriptors, functional status, or clinically grounded postoperative progression signals.

TABLE V. ABLATION: STRUCTURED-ONLY, UNSTRUCTURED-ONLY, AND UNSTRUCTURED AND AUDITED

| Model | ROC-AUC | P_1 | R_1 | F1 | Brier Score |
|---|---|---|---|---|---|
| Structured-only (LightGBM) | 0.95 | 0.69 | 0.62 | 0.65 | 0.04 |
| Unstructured-only (BioClinicalBERT, unaudited) | 0.88 | 0.61 | 0.68 | 0.64 | 0.09 |
| Unstructured-only (BioClinicalBERT, audited) | 0.85 | 0.66 | 0.57 | 0.61 | 0.07 |

Following attribution-guided lexical auditing and retraining, the audited model exhibited a reproducible redistribution of attribution patterns. Discharge-proxy tokens showed marked attenuation, and influence shifted toward broader postoperative context, including functional status descriptors, pain and symptom narratives, and progression-related language. Because auditing was the only intervention applied between these models, these attribution shifts support the interpretation that interpretability-guided masking can function as a training-time control mechanism, actively constraining what the model is permitted to learn without altering architecture or increasing supervision.

Quantitatively, auditing was associated with modest reductions in discrimination metrics, including ROC-AUC, accompanied by improved probability calibration reflected by lower Brier scores and more stable reliability behavior. Audited models produced fewer highly confident positive predictions near the decision boundary, consistent with a shift toward conservative behavior. In discharge planning contexts, where overconfident false-positive discharge recommendations can introduce operational disruption and safety risks, this conservatism represents a governance-aligned behavioral change that cannot be captured by discrimination metrics alone.

*F. Ablation Analysis of Structured and Unstructured Inputs*

Finally, we performed an ablation analysis to characterize the relative contribution and deployment properties of structured and unstructured modalities under strict temporal filtering and governance constraints. Three configurations were evaluated: structured-only prediction, unstructured-only prediction without auditing, and unstructured-only prediction with interpretability-guided auditing. Structured-only LightGBM demonstrated stable discrimination and favorable calibration behavior, consistent with the temporal stability of coded perioperative variables. Unstructured-only models showed substantially different behavior before auditing, including stronger dependence on discharge-related lexical shortcuts and degraded calibration. After auditing, unstructured-only models demonstrated reduced reliance on leakage-prone language and more conservative predictive behavior, with improved probability validity but reduced discrimination.

Performance comparisons across these configurations are summarized in Table V. While audited unstructured models exhibited modest decreases in ROC-AUC relative to unaudited unstructured counterparts, they consistently demonstrated improved calibration, as reflected by lower Brier score. These results show that unstructured clinical text provides complementary contextual signal but requires explicit governance to ensure temporal validity, and interpretability-guided auditing can partially align note-based behavior with the stability and calibration advantages observed in structured-only models.

V. DISCUSSION

This study reveals a deployment-relevant gap in how clinical NLP models for next-day discharge prediction are commonly evaluated. In the baseline early-fusion setting, gradient boosting achieved meaningful detection of next-day discharge, suggesting that temporally filtered note representations can complement perioperative structured variables for operationally useful risk stratification. However, once interpretability-guided lexical control was applied through an attribution-derived auditing mask, downstream classifiers shifted into a markedly conservative regime, most prominently at the default operating threshold (0.5). In this setting, predicted probabilities often remained below the actionable region even when rank-order separation persisted, highlighting that deployment utility depends on operating behavior at clinically meaningful thresholds, not only on discrimination metrics. The auditing mask was derived only from the training split via token-level SHAP aggregation and applied as a fixed lexical constraint before downstream training, ensuring that observed behavior reflects a controlled governance intervention rather than test-set contamination [10], [12].

The main contribution is not simply a reduction in retrospective performance. Auditing induced a measurable behavioral shift in how models expressed confidence and crossed decision thresholds. This divergence underscores why deployment-facing evaluation must consider not only ROC-AUC but also threshold-dependent actionability. In discharge planning workflows, clinical value arises when predictions reliably trigger actionable alerts under defined operating policies. When models rarely emit positive recommendations at the default threshold, they may preserve ranking signal while failing to support proactive coordination in practice.

Calibration trends must be interpreted alongside operating behavior. Auditing reduced global probability error, reflected by lower Brier scores and improved alignment of reliability curves, and attenuated highly confident positive predictions near the decision boundary. Nevertheless, improved probability validity does not guarantee clinical utility when calibrated probabilities remain concentrated below the actionable range under a fixed default threshold. This behavior suggests a mismatch between improved probability validity and a fixed default operating threshold, motivating threshold selection or stratified decision policies for deployment. From a safety perspective, reduced false-positive discharge recommendations may limit premature transitions and downstream workflow disruption, but this comes at the cost of missed opportunities for earlier planning and resource allocation [9], [16].

Several limitations define the scope of these conclusions. The auditing lexicon targeted a narrow set of explicit discharge-proxy terms and may not capture subtler documentation artifacts. In addition, the intervention operated through lexical suppression in the text stream, without directly auditing post-hoc attribution behavior of the fused classifier itself. Future work should expand auditing beyond explicit discharge terminology and evaluate decision-policy adaptation under governance constraints. The goal is not to maximize retrospective discrimination, but to prevent shortcut learning that undermines real-world decision support. Together, these findings support deploying discharge prediction as probability-based decision support with explicit operating policies (e.g., risk stratification and threshold tuning) while preserving clinician final authority.

## VI. Conclusions

We propose a lightweight interpretability-guided auditing pipeline that uses token-level attribution as a training-time control for clinical NLP under temporal leakage constraints. By integrating attribution-based screening into the modeling workflow, the approach limits reliance on documentation artifacts and promotes temporally valid signals. Across experiments, auditing reshaped attribution patterns and shifted model behavior toward more conservative, better-calibrated probability estimates without changing model architecture. These findings support interpretability-guided auditing as a practical governance strategy for deployment-oriented clinical NLP systems.